\documentclass[twocolumn, natbib]{svjour3}

\RequirePackage{fix-cm}
\usepackage{graphicx}%
\usepackage{multirow}%
\usepackage{amsmath,amssymb,amsfonts}%
\usepackage{mathrsfs}%
\usepackage[title]{appendix}%
\usepackage[pagebackref=true,breaklinks=true,citecolor=blue,linkcolor=blue,colorlinks]{hyperref}
\usepackage{xcolor}%
\usepackage{textcomp}%
\usepackage{manyfoot}%
\usepackage{booktabs}%
\usepackage{algorithm}%
\usepackage{algorithmicx}%
\usepackage{algpseudocode}%
\usepackage{listings}%
\smartqed
\usepackage{bm}  
\usepackage{xspace} 
\usepackage{booktabs}
\usepackage{adjustbox}
\usepackage{makecell}
\usepackage{mwe}
\usepackage{ulem}
\usepackage{subcaption}

\newcolumntype{R}[2]{%
    >{\adjustbox{angle=#1,lap=1.3\width-(#2)}\bgroup}%
    l%
    <{\egroup}%
}

\begin{document}\sloppy

\newcommand{\systemname}{{HVDistill}\xspace}
\newcommand{\parsection}[1]{\vspace{4pt}\noindent\textbf{#1:}}
\newcommand{\psection}[1]{\vspace{4pt}\noindent\textbf{#1\\}}
\definecolor{mgreen}{RGB}{1,150,74}
\newcommand\up[1]{\textcolor{mgreen}{$^{\uparrow{#1}}$}}

\title{HVDistill: Transferring Knowledge from Images to Point Clouds via Unsupervised Hybrid-View Distillation}

\author{Sha Zhang$^{1,3}$ \and Jiajun Deng$^2$ \and Lei Bai$^3$ \and Houqiang Li$^1$ \and Wanli Ouyang$^3$ \and Yanyong Zhang$^1$}
\authorrunning{Sha Zhang~\etal} 

\institute{
	Sha Zhang \at
	\email{zhsh1@mail.ustc.edu.cn}           
	\and
	Jiajun Deng, corresponding author \at
	\email{jiajun.deng@adelaide.edu.au}
    \and
    Lei Bai \at
	\email{baisanshi@gmail.com}
	\and
	Houqiang Li \at
	\email{lihq@ustc.edu.cn}
	\and
	Wanli Ouyang \at
	\email{wanli.ouyang@sydney.edu.au}
	\and
	Yanyong Zhang, corresponding author \at
	\email{yanyongz@ustc.edu.cn}
	\\
 \\
$^1$University of Science and Technology of China, Hefei, Anhui, China\\
\\
$^2$The University of Adelaide, Adelaide, Australia\\
\\
$^3$Shanghai AI Laboratory, Shanghai, China
}

\date{Received: date / Accepted: date}

\maketitle
\begin{abstract}
We present a hybrid-view-based knowledge distillation framework, termed HVDistill, to guide the feature learning of a point cloud neural network with a pre-trained image network in an unsupervised manner. By exploiting the geometric relationship between RGB cameras and LiDAR sensors, the correspondence between the two modalities based on both image-plane view and bird-eye view can be established, which facilitates representation learning. Specifically, the image-plane correspondences can be simply obtained by projecting the point clouds, while the bird-eye-view correspondences can be achieved by lifting pixels to the 3D space with the predicted depths under the supervision of projected point clouds. The image teacher networks provide rich semantics from the image-plane view and meanwhile acquire geometric information from the bird-eye view. Indeed, image features from the two views naturally complement each other and together can ameliorate the learned feature representation of the point cloud student networks. Moreover, with a self-supervised pre-trained 2D network, HVDistill requires neither 2D nor 3D annotations. We pre-train our model on nuScenes dataset and transfer it to several downstream tasks on nuScenes, SemanticKITTI, and KITTI datasets for evaluation. Extensive experimental results show that our method achieves consistent improvements over the baseline trained from scratch and significantly outperforms the existing schemes. Codes are available at git@github.com:zhangsha1024/HVDistill.git.

\end{abstract}

\section{Introduction}
\label{sec:intro}

\def\svgwidth{\linewidth}
\begin{figure}[t]
  \centering
  \includegraphics[width=1.0\linewidth]{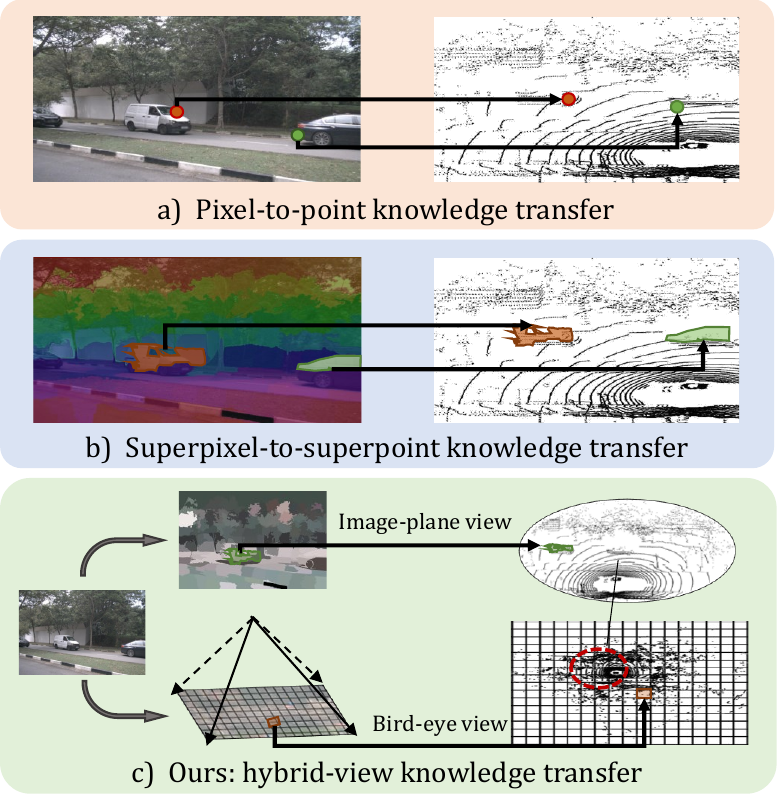}
	\caption{Comparison between existing schemes and ours. Both (a)  and (b)  transfer image knowledge to point cloud networks based on the image-plane view. In contrast, we develop a hybrid-view framework (c) to transfer image knowledge based on both the image-plane view and the bird-eye view.  }
	\label{fig:intro-compare}
\end{figure}
\def\svgwidth{\linewidth}

LiDAR point clouds play a vital role in 3D perception for numerous applications, such as autonomous driving~\citep{zhu2022vpfnet,li2022mathsf,wang2023multi},  virtual reality~\citep{han2021framework, alexiou2020pointxr}, and domestic robotics~\citep{qi2020building,  duan2022pfilter}, etc. However, collecting and annotating point clouds incur prohibitive costs~\citep{behley2019semantickitti, caesar2020nuscenes}, limiting both the quantity and quality of the available datasets to date. The pretrain-and-finetune paradigm, which has achieved inspiring successes on 2D images, cannot be directly applied to the 3D domain. As a result, the previously developed point cloud processing models are generally initialized from scratch and trained with an enormous amount of samples~\citep{shi2023pv}. The absence of a properly pretrained model has thus become one of the major stumbling blocks in the 3D perception field~\citep{xie2020pointcontrast, liu2021learning, sautier2022image}. 

In fact, neural networks can learn from unsupervised pretext tasks with a carefully designed loss function~\citep{he2022masked, grill2020bootstrap, he2020momentum}, which is known as unsupervised feature learning (UFL). In 2D image domain, UFL has been validated to be extremely effective in providing powerful and transferable feature representation for various downstream tasks~\citep{xiao2022learning, xie2022delving}. However, when it comes to 3D point clouds, UFL hasn't shown such impact. The main reason is that the success in images is based on millions or even billions of training samples, while datasets at such scale are unavailable for point clouds. Besides, compared to image pixels, LiDAR points are sparse and unevenly distributed, rendering it harder to learn the appropriate feature representations.

Alternatively, the properly calibrated images and point clouds that are often available for autonomous vehicles can provide a potential solution to transfer knowledge learned from images to point clouds in a cross-modality manner. As shown in Figure~\ref{fig:intro-compare}(a), the early work PPKT~\citep{liu2021learning} performs 2D-to-3D knowledge transfer via pixel-to-point contrastive learning. However, the pixel-to-point correspondence needed by this approach is imperfect in autonomous driving scenes due to issues such as sensor occlusion and motion blur. To address this problem, the more recent work SLidR~\citep{sautier2022image} proposes to use superpixels and superpoints instead for more reliable cross-modal correspondences. As illustrated in Figure~\ref{fig:intro-compare}(b), SLidR first groups visually similar pixels into superpixels, and then performs 2D-to-3D knowledge transfer via contrastive loss between superpixels and superpoints. Here, a superpoint indicates a set of points projected to the same superpixel.

These prior studies attempt to learn point cloud representation from an image teacher network by developing 2D-3D correspondences based on the image-plane view (IPV). However, images are inherently insufficient to represent 3D scenes due to the unavailability of 3D geometric information. As such, the 2D features extracted from the IPV alone naturally fall short in guiding the representation learning of a point cloud network, taking risks in ignoring 3D spatial layouts. 
In this paper, we strive to enhance the 2D-to-3D knowledge transfer by also considering the 2D features from other views that can be derived from images.  Formally, we introduce a \emph{hybrid-view} framework, termed HVDistill, to distill image knowledge to point cloud networks in an unsupervised manner. As shown in Figure~\ref{fig:intro-compare}(c), 
In this paper, we strive to enhance the 2D-to-3D knowledge transfer by also considering the 2D features from other views that can be derived from images.  Formally, we introduce a \emph{hybrid-view} framework, termed HVDistill, to distill image knowledge to point cloud networks in an unsupervised manner. As shown in Figure~\ref{fig:intro-compare}(c), 
in addition to IPV, our HVDistill further leverages the bird-eye view(BEV) as a complement for knowledge transferring. As such, the usage of both the image-plane view and bird-eye view yields a virtual 3D space, from which better 3D representations can be learned, with both semantic and geometric information taken into consideration.

The most notable design consideration of HVDistill is the addition of the image's BEV features. Specifically, when projecting 3D points to the image plane as in previous methods~\citep{liu2021learning, sautier2022image}, we not only build the correspondences between points and pixels but also obtain the depth information for the pixels with projected points. 
However, due to the sparsity of point clouds, only a small fraction of pixels has direct correspondences with points, which means huge potential untapped. In our algorithm, we address this issue by estimating a dense depth map for each image under the supervision of the sparse depth map directly provided by point clouds. We combine the image features with camera intrinsic to estimate the depth for each pixel. Once obtaining the depth, the pixels can be ``lifted'' to the 3D space~\citep{wang2019pseudo}.
 Next, we splat the lifted pixels to the BEV plane, because the BEV representation has been validated to be effective in 3D perception tasks for autonomous driving~\citep{liu2022bevfusion, huang2021bevdet, li2022bevformer}, where the objects are not likely to overlap with each other and their sizes are consistent with the real world ignoring the distance to the ego sensor. 
 
To validate the merits of our proposed framework, we first conduct pre-training on nuScenes dataset\citep{caesar2020nuscenes}, and then evaluate the learned representation on two typical 3D perception tasks, \emph{i.e.}, 3D semantic segmentation and 3D object detection, over three prevalent benchmarks, \emph{i.e.}, nuScenes-Lidarseg\citep{caesar2020nuscenes}, SemanticKITTI\citep{behley2019semantickitti} and KITTI\citep{geiger2012we}. Our HVDistill achieves improvements on all three datasets. Remarkably, the model pretrained with HVDistill achieves 49.7\% mIoU, resulting in 5.1\%  performance improvement over the strongest counterpart for few-shot semantic segmentation on SemanticKITTI, and up to 8.7\% mAP improvements for few-shot object detection.

In summary, our main contribution is the HVDistill framework that transfers image knowledge to point cloud networks via cross-modality contrastive distillation based on image-plane and bird-eye views. Compared to prior works, the hybrid-view image teachers in our HVDistill take both semantic and geometric information into account, thus learning the effective representation from point clouds. Moreover, we provide an elegant solution to involve the image's BEV features with marginal cost and no extra annotations. Our HVDistill takes a strong step towards effective pre-trained networks for 3D point clouds and exhibits great potential to serve as a baseline for future investigation.

\def\svgwidth{\linewidth}
\begin{figure*}[t]
  \centering
  \includegraphics[width=1.0\linewidth]{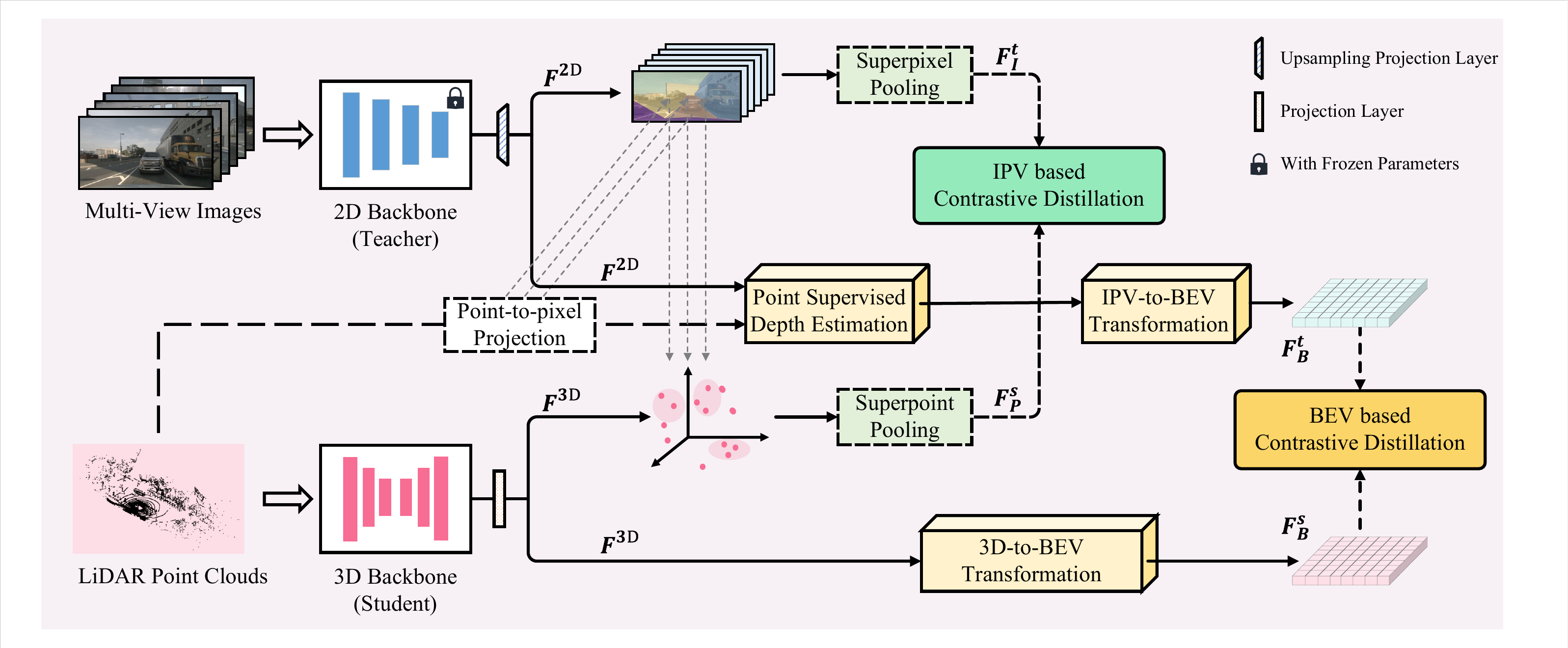}
	\caption{\textcolor{black}{\textbf{An overview of the proposed \systemname pipeline.} Our approach transfers image knowledge from a pre-trained 2D network into a 3D neural network via hybrid-view contrastive distillation. 
	On one hand, the point clouds are grouped into superpoints according to the corresponding superpixels generated on each image, and then supervised by the image features from the 2D teacher network by image-plane view (IPV) based contrastive distillation.
	On the other hand, the features of images/point clouds from  2D/3D backbones are transformed to the bird-eye view (BEV), and then the image BEV features are used for supervising the point cloud BEV features by contrastive loss.
    Note that the 2D backbone's parameters are frozen.}  
    }
	\label{fig:overview}
\end{figure*}

\def\svgwidth{\linewidth}

\section{Related Work}
\label{sec:related}
In this section, we briefly review unsupervised feature learning (UFL) and cross-modality knowledge distillation.

\subsection{UFL for Point Clouds}
    Unsupervised feature learning aims to learn general feature representation without human-annotated labels. The learned representation is supposed to be easily adapted to downstream tasks via fine-tuning. 

    The existing intra-modality UFL methods for point clouds can be  divided into two categories: generative-based methods and contrastive-based methods. 
    
    Generative-based methods~\citep{han2019view,han2021hierarchical,chen2021unsupervised,han2019multi,rao2020global,chen2019deep,zhao20223dpointcaps++}, %yang2017foldingnet} 
    explore 3D data reconstruction as pretext tasks to learn effective feature representation. Typically, a series of pretext tasks are involved, including 3D objects reconstruction\citep{chen2021unsupervised,han2019multi,zhao20223dpointcaps++},  point clouds completion\citep{rao2020global,chen2019deep,yang2017foldingnet} and rotation prediction\citep{han2019view,han2021hierarchical}. Besides, PC-GAN\citep{li2018point} introduces generative adversarial networks to the point clouds domain, learning point clouds' feature via joint distribution estimation.     
    
    Contrastive-based methods~\citep{xie2020pointcontrast,zhang2021self,wang2021unsupervised,jiang2021unsupervised} follow the paradigm to learn effective point clouds features by encouraging the augmentations of the same input to approach each other while pushing away that of different inputs. 
    Info3D~\citep{sanghi2020info3d} proposes to maximize the mutual information between the whole 3D objects' point clouds and their chunks to improve the feature representation. PointContrast\citep{xie2020pointcontrast}, for the first time, generalizes self-supervised representation learning to point clouds in complex scenes by leveraging the correspondence between points from different camera views and shows that the pre-trained models can be efficiently transferred to facilitate multiple downstream tasks (e.g., classification, object detection, part segmentation, and semantic segmentation). Unlike PointContrast which relies on correspondences between points from different camera views, DepthContrast~\citep{zhang2021self} shows that performing contrastive learning with the global feature of the whole 3D scene is enough to learn effective representation.

    \subsection{Cross-Modality Knowledge Distillation}

    Knowledge distillation (KD)~\citep{buciluǎ2006model,hinton2015distilling} is first introduced to train a small model (\emph{i.e.}, student)
    with the guidance of a well-performing large one (\emph{i.e.}, teacher) for model compression. The follow-up works~\citep{mirzadeh2020improved,cho2019efficacy,zhang2020improve,zhao2022decoupled} explore KD in various tasks and demonstrate its effectiveness in knowledge transferring,  especially via a cross-modality manner~\citep{gupta2016cross,guo2021liga,alwassel2020self,tian2019contrastive,liu20213d}.

    Cross-modality KD has attracted increasing attention in 3D perception fields. Among the surge of research, several recent works~\citep{liu2021learning,sautier2022image} distill knowledge in the same setting as ours: treating a well-pretrained image model as the teacher and a randomly initialized point clouds model as the student. Specifically,  PPKT~\citep{liu2021learning} optimizes the point cloud features network by mimicking the corresponding image features in a pixel-to-point manner. However, the accurate matching of pixel-to-point is not always accessible in real-world data, especially in autonomous driving, hampering the effectiveness of the learned point cloud feature. Instead, SLidR~\citep{sautier2022image} proposes to use superpixels-to-superpoints as an alternative. It groups visually similar pixels to superpixels and obtains corresponding superpoints by projection.  The contrastive distillation is then performed among superpixels and superpoints.
     
    From our point of view, these two works both conduct contrastive learning based on the correspondence of image-plane view, which is insufficient for point cloud representation learning. In contrast, we argue that BEV-level features can provide a different viewpoint and compensates for the 3D geometric information. Meanwhile, the combination of image-plane view and BEV can narrow the domain gap during 2D-to-3D distillation, achieving more excellent performance.

\section{Methodology}
\systemname is an unsupervised pre-training pipeline that transfers knowledge from a pre-trained image network to a point cloud network via hybrid-view knowledge distillation.

An overview of \systemname is depicted in  Figure~\ref{fig:overview}.
Given paired point clouds and images, a 3D backbone network and a 2D backbone network are applied to extract 3D point cloud features and 2D image features, respectively.
After that, knowledge transferring is conducted through the image-plane view (IPV) based and bird-eye view (BEV) based contrastive distillation. On the one hand, pixels are grouped into superpixels, and superpoints are generated accordingly. The feature of superpixel is obtained by average pooling over image features, and the feature of superpoint is obtained by average pooling over point cloud features. Then, IPV-based contrastive distillation is performed based on superpixels and superpoints.
On the other hand,
we transform image features from IPV to BEV in three steps: (1) we project points to the image plane by geometric relationship to generate a sparse depth map with accurate depth value; 
(2) with the supervision of the sparse depth map, we predict a dense depth map via a depth prediction head; and (3) we lift the pixel-level features to the 3D space according to the depth and then collapse it to BEV to produce image BEV features.

The image BEV features are leveraged to supervise the point clouds BEV features by grid-to-grid BEV-based contrastive distillation.

In the following subsections, we first briefly introduce preliminary information and then present the detailed design of our hybrid-view distillation algorithm.

\subsection{Preliminary}

Before introducing the hybrid-view distillation framework and training objectives of \systemname, we briefly review knowledge distillation (KD) and contrastive learning (CL) techniques, which serve as the preliminary of our work.

\parsection{Knowledge Distillation (KD)}
The conventional target of KD is to transfer the knowledge from an unwieldy network to a smaller network. The network to provide knowledge is known as a teacher network, and the network to be learned is the student network. In this work, we follow the recent practices~\citep{gupta2016cross,guo2021liga,liu20213d} to extend this conventional target to a general one. Particularly, HVDistill is intended to transfer the knowledge of 2D networks trained with images to a 3D network aiming at point clouds, learning the benefits of rich semantic information from images. We denote the 2D teacher network as $\mathcal{N}^t(\cdot)$ and the 3D student network as $\mathcal{N}^s(\cdot)$. The parameters of $\mathcal{N}^t(\cdot)$ are frozen in our method.

\parsection{Contrastive Learning (CL)} 
CL is widely adopted as a pretext task for unsupervised representation learning~\citep{he2020momentum,wang2021unsupervised,jiang2021unsupervised}, with the  training objective to reduce the feature distance between positive pairs while increasing that between negative ones.
Since the traditional margin-based loss function is hard to be optimized, the CL methods usually use the noise contrastive estimation loss InfoNCE~\citep{oord2018representation}, which is computed as follows:
\begin{equation}
\mathcal{L}_\text{InfoNCE} = - \log \frac{ exp ( (F_k)^T F'_k /\tau) }{ \sum_{i=1 }^{N} exp ((F_k)^T F'_i /\tau)}. 
\end{equation}

Our \systemname follows the feature-based paradigm to perform images to point clouds KD. Knowledge is transferred via image plane view and bird-eye view with two InfoNCE loss variants in an unsupervised manner. 

\subsection{Hybrid-View Distillation}
The primary innovation of this work is the hybrid-view contrastive distillation for point cloud network pre-training. Given a scan of point cloud ${P}_0  \in \mathbb{R}^{N\times 3}$ and multi-view images $\mathcal{I} = \{I_1, I_2,...,I_K\ |\ I_i \in \mathbb{R}^{H \times W \times 3} \} $ as the inputs, our method first applies the 3D/2D backbones to extract point/image features. A high-resolution feature map is necessary for establishing point-to-pixel correspondences, while the output of the 2D backbone network (\emph{i.e.}, ResNet) is usually overly downsampled for the correspondence. We replace the downsampling convolutional layers with dilation convolutional layers following previous methods\citep{sautier2022image} and add an upsampling projection layer $\mathcal{H}^t$ following the 2D backbone $\mathcal{N}^t(\cdot)$ to obtain a high-resolution image feature map $\bm{F}^\text{2D}_k \in \mathbb{R}^{H \times W \times C}$, where k indicates the $k$-th view. Note that the spatial resolution of $\bm{F}^\text{2D}_k$ is the same as that of input images but with $C$ channels.
Another projection layer $\mathcal{H}^s$ is then applied to the point features out of the 3D backbone $\mathcal{N}^s(\cdot)$ to produce point cloud features $\bm{F}^\text{3D}\in \mathbb{R}^{N \times C}$ that match the feature dimension of each $\bm{F}^\text{2D}_k$. Our proposed hybrid-view distillation, including IPV-based contrastive distillation and BEV-based contrastive distillation, is built on $\{\bm{F}^\text{2D}_k\}_{k=1}^K$ and $\bm{F}^\text{3D}$.

\parsection{IPV based Contrastive Distillation}
IPV-based contrastive distillation transfers knowledge from 2D backbone to 3D backbone via cross-modality contrastive learning between 2D features from the image plane and point cloud features from the 3D space. We name it IPV based since the teacher signals, \emph{i.e.}, 2D features, are from the image plane. 

Inspired by the method in~\citep{sautier2022image}, we perform contrastive learning at the cluster level. Specifically, visually similar pixels in each image are grouped together to generate superpixels with a bottom-up segmentation algorithm, SLIC\citep{achanta2012slic}. We use $\bm{C}^m_k$ to denote the set of pixels belonging to the $m$-th superpixel from the $k$-th image. An L2 normalization function followed by an average pooling is applied on $\bm{F}^\text{2D}$ to produce the superpixel's feature $\bm{F}^{t}_{I(k,m)}$ as:
\begin{equation}
\bm{F}^{t}_{I(k,m)} =\dfrac{1}{\left| C^m_k \right| } \sum_{(x,y) \in C_k^m }\frac{\bm{F}_k^\text{2D}(x,y)}{||\bm{F}_k^\text{2D}(x,y)||_2},
\end{equation}
where $(x,y)$ is the index of a pixel in the image feature map.
In the 3D space, we use superpoints as a cluster to learn from superpixels. \textcolor{black}{To generate superpoints, we first obtain the point-pixel correspondence by projecting the point clouds to the image plane according to the extrinsic $T_{\mathrm{l2c}}$ and intrinsic matrices $T_{\mathrm{c2i}}$. Once the point- pixel correspondence is acquired, the points projected to the same superpixel are grouped together to formulate a superpoint $D_k^m$:
  \begin{equation}
 	D_k^m=\mathcal{G}(T_{\mathrm{c2i}} T_{\mathrm{l2c}} \mathcal{P}, C_k^m)
\end{equation}}
Note that the correspondence between points and pixels is independently obtained for each image, thus superpoints are different for different image views. Similar to producing features of superpixels, L2 Normalization, and an average pooling layer are applied to compute the superpixel's feature:
\begin{equation}
F^{s}_{P(k,m)} =\dfrac{1}{\left| D^m_k \right| } \sum_{j \in D_k^m }\frac{\bm{F}^\text{3D}(j)}{||\bm{F}^\text{3D}(j)||_2},
\end{equation}
where $j$ is the index of a point from the 3D features. 

After obtaining features of superpixels and superpoints, we utilize superpixel-to-superpoint contrastive loss to perform point cloud feature learning. The matched superpoint-superpixel pairs are taken as positive samples, while unmatched ones are token as negative samples. In an ideal situation, the positive samples are close to each other, while the negative samples are far away from each other. Specifically, the superpixel-to-superpoint contrastive loss is adapted based on InfoNCE loss~\citep{oord2018representation}, and calculated as: 
\begin{equation}
\begin{aligned}
&\mathcal{L}_\text{IPV} = - \sum_{k,m} \log \left[ \frac{ E_{k,m}(k,m) }{ \sum_{k',m'} E_{k,m}(k',m')} \right],\\
& \text{with} \  E_{k,m}(k',m')=exp (\ (\bm{F}^{t}_{I{(k,m)}})^T \bm{F}^{s}_{P{(k',m')}} \ /\tau),
\end{aligned}
\end{equation}
where $\tau > 0$ is a temperature coefficient that is used to smooth the sample distribution. In this situation, the objective is to minimize the distance between the feature $F^{t}_{I{(k,m)}}$ of a superpoint and the feature $F^{s}_{P{(k,m)}}$ of the corresponding superpixel within the same image and superpixel region. At the same time, the aim is to maximize the distance between superpixel features and superpoint features that do not share the same indices or belong to different scenes.

\parsection{BEV based Contrastive Distillation}
BEV-based contrastive distillation provides an additional view for knowledge transfer.  By transforming image features and point cloud features to BEV, a complementary pathway is created to enhance the transfer of knowledge.
With the supervision of point clouds, the synthesized image BEV features can effectively preserve the geometric properties and thus ameliorate the feature learning of the 3D student network.
There are two steps to obtain the image BEV features: (i) point supervised depth estimation, and (ii) IPV-to-BEV transformation.

In the first step, we reuse the correspondences established from point-to-pixel projection to obtain the depths of those pixels that have matching points. Since the LiDAR point clouds are far fewer than image pixels, the obtained depth map $\bm{D}_\text{sparse} \in \mathbb{R}^{H\times W\times 1}$ is too sparse to lift the 2D image features into 3D space. 

To address this issue, we add a depth prediction head in our point-supervised depth estimation module to output a dense depth map $\bm{D}_\text{dense}\in\mathbb{R}^{H_0\times W_0\times T}$, each element of which represents $T$ discrete depth ranges. 
\textcolor{black}{Specifically, the sparse depth map $\bm{D}_\text{sparse}$ generated from point clouds is leveraged as the depth supervision.} Moreover, to make the depth camera aware, we combine camera intrinsic matrices $\bm{M}$ with image features to form the input of the depth prediction head. The computation of $\bm{D}_\text{dense}$ can be described as:
\begin{equation}
\bm{D}_\text{dense} = {\mathcal{N}_D(SE(MLP(\bm{M}),Conv(\bm{F}^\text{2D})))},
\end{equation}
where $\mathcal{N}_D(\cdot)$ is the depth prediction head, $SE(\cdot)$ indicates a Squeeze-and-Excitation module, %$(\cdot,\cdot)$ means concatenation, 
and $Conv(\cdot)$ represents linear transformation with a convolutional layer. The depth prediction head consists of three Residual Blocks and a Deformable Convolution layer. We use Binary Cross Entropy for the depth loss $\mathcal{L}_\text{depth}$.

In the subsequent step, we lift the image features to 3D space by expanding the pixel-level features along the direction of camera rays and weighting each point feature with the corresponding response from $\bm{D}_\text{dense}$. We then group the lifted 3D features within each $r \times r$ grid in the X-Y plane together and collapse them to the BEV along the Z-axis by summation. Finally, we use an L2 Normalization layer to generate the final image BEV feature map $\bm{F}^{t}_{B}\in \mathbb{R}^{H_b \times W_b \times E}$. This enables us to lift the image features to 3D and generate a BEV feature map, which is a key step in our approach.

The view transformation of point cloud features from the 3D space to the BEV is more straightforward.
We simply flatten the 3D point cloud features $\bm{F}^\text{3D}$ by reshaping the features along Z-axis to the channel dimension. After that, three extra convolutional layers followed by an L2 Normalization layer are applied to obtain the point cloud BEV features $\bm{F}^{s}_{B}\in \mathbb{R}^{H_b\times W_b \times E}$.

The image BEV feature map $\bm{F}^t_B$ and the point cloud feature map $\bm{F}^s_B$ are in the unified BEV representation with the same resolution, which is naturally aligned. The feature embedding $\bm{F}^{s}_B{(i,j)}$ from point cloud BEV features is supposed to be similar to the corresponding grid feature $\bm{F}^{t}_B{(i,j)}$ from the image BEV features, and quite different from the grid features in other positions or other scenes. Based on this principle, we devise the training objective of our BEV-based contrastive distillation as:
\begin{equation}
\begin{aligned}
& \mathcal{L}_\text{BEV} = - \sum_{i,j} \log \left[ \frac{ E'_{i,j}(i,j) }{ \sum_{i',j'} E'_{i,j}(i',j')} \right],\\
& \text{with} \quad \  E'_{i,j}(i',j')=exp (\ (\bm{F}^{t}_{B{(i,j)}})^T \bm{F}^{s}_{B{(i',j')}} \ /\tau),
\end{aligned}
\end{equation}
where $\tau' > 0$ is a temperature factor. In other words, if an image BEV grid and a point BEV grid have the same index in the BEV map for the same scene, they are positive pairs and are supposed to be close to each other in the feature space. Otherwise, they are negative pairs and are supposed to be far away from each other. In practice, we only take non-zero grids in point cloud BEV maps for training to avoid the gradient collapse problem caused by empty grids.

\parsection{Overall Loss}
By combining the training objectives from the IPV based, BEV based pathways, and the depth prediction module, we get the loss function:
 % overall
\begin{equation}
\mathcal{L} = \alpha \mathcal{L}_\text{IPV} +\beta \mathcal{L}_\text{BEV} +\gamma \mathcal{L}_\text{depth},
\end{equation}
where $\alpha, \beta, \gamma$ are the weights of each training objective.

\section{Experiments}
We evaluate \systemname pre-training method by fine-tuning and linear-probing on downstream tasks over several datasets. For pre-training, we choose a public large-scale multi-modality autonomous driving dataset, i.e., nuScenes~\citep{caesar2020nuscenes}. Then we fine-tune and evaluate the pre-trained model on nuScenes, nuScenes-lidarseg~\citep{caesar2020nuscenes}, SemanticKITTI~\citep{behley2019semantickitti}, and KITTI~\citep{geiger2012we} for different downstream tasks separately. In this section, we present the implementation details in parts: 1) datasets and metrics used in pre-training and fine-tuning. 2) the network architecture. 3) the pre-training details. 4) the downstream experiments, including transferring to point clouds semantic segmentation and few-shot 3D object detection. 5) ablation study to validate the superiority of the design.

\subsection{Datasets and Metrics}

\parsection{nuScenes } 
We conduct all the pre-train experiments on this dataset. nuScenes contains 1000 scenes with 150 for testing, 150 for validation, and others for training. The whole dataset includes approximately 1.4M camera images and 390k LIDAR sweeps. Each scene has around 40 keyframes. Every keyframe provides one scan of point cloud from the LIDAR, six images from six different cameras, and the corresponding synchronization and calibration information. 

\parsection{nuScenes-lidarseg } 
nuScenes-lidarseg is an extension for nuScenes. This dataset has semantic labels of 32 categories and annotates each point from keyframes in nuScenes. We use the 700 scenes in the training set with segmentation labels to fine-tune for the semantic segmentation task, and the 150 scenes in the validation set to verify the performance.

\parsection{SemanticKITTI }
There are 28 semantic classes in SemanticKITTI, and 19 are calculated for evaluation. The annotation cover traffic participants and ground class such as sidewalks. SemanticKITTI contains 22 sequences with only 00-10 annotated. We use the 8th sequence of SemanticKITTI to evaluate the quality of semantic segmentation task and the other 10 sequences for fine-tuning.

We fine-tune our pre-trained model transferring on segmentation on SemanticKITTI, which never shows in the pre-training stage. 
  
\parsection{KITTI }
KITTI is a classical dataset for 3D object detection. It  contains 7481 training point clouds and 7518 test point clouds. We follow the official development kit to partition them as training set and validation set.  KITTI only evaluates three kinds of objects, including cars, cyclists, and pedestrians. We conduct object detection experiments on the training set of KITTI object detection dataset.

\parsection{Evaluation Metrics } 
For fine-tuning on semantic segmentation, we report mIoU and fwIoU validated in specific data. For fine-tuning on object detection, we follow the KITTI\citep{geiger2012we} comparing methods by mAP: First, we get the AP\_R40 (40 recall positions)  at an overlap of 0.7 for cars, 0.5 for pedestrians, and 0.5 for cyclists. Then, we compute the average AP over these three classes in the moderately difficult cases. 

\subsection{Network Architectures}

\parsection{Image Teacher Network}
We use a ResNet-50\citep{he2016deep} as our image teacher network. This backbone network is pre-trained with MoCov2\citep{chen2020improved,he2020momentum} on ImageNet, enabling our whole training process to get rid of annotations. To maintain the receptive field without reducing the spatial resolution, the second and following stridden convolutions are replaced with dilated convolutions as the same in previous methods\citep{sautier2022image}. The adoption of the dilation strategy enlarges the resolution of the output feature map from $\frac{1}{32}$ to $\frac{1}{4}$ of the input images. Besides, with the upsampling projection layer, the resolution is further recovered to the same as that of input images.

\parsection{Point Cloud Student Network}
Following the common practice as in \citep{xie2020pointcontrast,zhang2021self,sautier2022image}, we adopt the SR-UNet~\citep{choy20194d} as our 3D backbone network, \emph{i.e.}, the student model. SR-UNet has 256 output channels, while image features from the projection layer have 64 channels. We use a fully connected layer after SR-UNet to match the point cloud feature channels with image feature channels.  
We quantize the 3D points to voxels as the input of SR-UNet. We generate voxels in Cartesian coordinates with the X-axis range of [-51.2m, 51.2m], Y-axis range of [-51.2m, 51.2m], and Z-axis range of [-5.0m, 3.0m]. The voxel size is set as (0.1m, 0.1m, 0.1m).

%---------------------table-1----------------------------------------------------
\begin{table*}
\setlength\tabcolsep{9pt}
  \centering
  \caption{\textcolor{black}{Performance comparison of different pre-training methods for semantic segmentation by fine-tuning. On nuScenes and SemanticKITTI we use only 1\% of the annotated training data, respectively. The results with * are from \citep{sautier2022image}. The results are the $mIoU (\%)$ on the validation sets of nuScenes and SemanticKITTI.} }%The best results are marked with bold font. }
  \addtolength\tabcolsep{6.5pt}
  \begin{tabular}{c|cl|cc}
    % \toprule
    \hline
    \multirow{2}{*}{Initialization} &  \multicolumn{2}{c|}{nuScenes\citep{caesar2020nuscenes}} & \multicolumn{2}{c}{SemanticKITTI\citep{behley2019semantickitti}} \\
    & mIoU & \textcolor{black}{P-gain} & mIoU & \textcolor{black}{P-gain} \\
    \hline
    Train from scratch & 28.3 &\_& 31.4&\_ \\
    PointContrast*\citep{xie2020pointcontrast} & 32.5&\small{+4.2} & 41.1&\small{+9.7}\\
    DepthContrast*\cite{zhang2021self} & 31.7&\small{+3.4} & 41.5&\small{+10.1} \\
    PPKT*\cite{liu2021learning} & 37.8&\small{+9.5} & 43.9&\small{+12.5} \\
    SLidR*\citep{sautier2022image} & 38.3&\small{+10.0} & 44.6&\small{+13.2} \\
    \textbf{\systemname (Ours)} & \textbf{42.7\up{4.4}}&\small{+14.4} & \textbf{49.7\up{5.1}}&\small{+18.3}  \\
    % \bottomrule
    \hline
  \end{tabular}
  \label{tab:segmentation}
\end{table*}

%------------------------------table-2-------------------------------------------
\begin{table*}
  \centering
  \caption{\textcolor{black}{Performance comparison of random initialization and our pre-trained backbone using our method \systemname for semantic segmentation with different percentages of annotated training data on nuScenes. The results are the $mIoU (\%)$ on the validation set of nuScenes.} } 
  \addtolength\tabcolsep{0pt}
  \resizebox{\linewidth}{!}
  {
  \begin{tabular}{c|cl|cl|cl|cl|cl}
    \hline
    \multirow{2}{*}{Initialization} & \multicolumn{2}{c|}{1\% } &  \multicolumn{2}{c|}{5\%} &  \multicolumn{2}{c|}{10\%}  &  \multicolumn{2}{c|}{25\%}  &  \multicolumn{2}{c}{100\%} \\
    & \small{mIoU} & \small{P-gain} & \small{mIoU} & \small{P-gain} & \small{mIoU} & \small{P-gain} & \small{mIoU} & \small{P-gain} & \small{mIoU} & \small{P-gain} \\
    \hline
    Train from scratch & 28.3 & \_ & 46.2 & \_ & 56.6& \_& 64.0& \_& 74.2& \_\\
    SLidR\citep{sautier2022image} & 38.3& \small{+10.0} & 52.2& \small{+6.0} & 58.8& \small{+2.2} & 66.2& \small{+1.8} & 74.6& \small{+0.4} \\
    {\textbf{\systemname(Ours)}} & {\textbf{42.7\up{4.4}} }& \small{+14.4} & \textbf{56.6\up{4.4}} & \small{+10.4}& \textbf{62.9\up{4.1}}& \small{+6.3} & \textbf{69.3\up{3.5}} & \small{+5.3}& \textbf{76.6\up{2.0}}& \small{+2.4}\\
    \hline
  \end{tabular}}
  \label{tab:percentagesegmentation}
  
\end{table*}

\parsection{Other Modules}
The upsampling projection layer $\mathcal{H}^t$ is composed of a $1 \times 1$ convolutional layer and an upsampling layer. 
The convolutional layer reduces the channels from 2048 to 64.
The upsampling layer performs bi-linear interpolation with a scale factor of 4 to adjust the solution of image features. The projection layer  $\mathcal{H}^s$ is a fully-connected layer to align the channels as 64. 
Besides, the discrete depth size $T$ is set as 118 and the BEV map size is $256\times256$.

\subsection{Pre-training Details}
\label{subsection:pretrain}
The 3D network backbone and all the learnable heads are pre-trained on 4 GPUs with a batch size of 16 for 50 epochs. We choose the SGD optimizer with a momentum of 0.9 and a weight decay of 0.0001. This optimizer is also used in all the training of downstream tasks. The initial learning rate is 0.5. A cosine annealing scheduler is employed to adjust the learning rate from the initial value to 0. The temperature hyperparameter in superpixel-to-superpoint contrastive loss and grid-to-grid contrastive loss are all set to 0.07. The coefficients $\alpha,\beta$ are set to 0.25 and 1, respectively.

\noindent\textbf{Data Augmentation:} We apply several augmentations in the pre-training of the backbone. For 3D point clouds, we use random rotation, flip, scale, translate, and drop points that lie in a random cuboid. On the 2D component, we crop and resize to get a patch with the size of [224, 416]. Besides, we apply a random horizontal flip, as in SLidR\citep{sautier2022image}. Note that we have to inverse those operations to align the image BEV feature and the point cloud BEV feature.

%-------------------------------------------------------------------------
\begin{table*}[t]
  \centering
  \caption{\textcolor{black}{Performance comparison of methods for object detection by fine-tuning pre-trained networks using different percentages of the annotated training data in the KITTI. We present the average $mAP (\%)$ of cars, cyclists, and pedestrians in moderately difficult cases.}} %The best results are marked with bold font. }
  \addtolength\tabcolsep{15pt}
  \begin{tabular}{c|c|c|c|c}
    \hline
    \small{Initialization} & 1\%& 5\%   & 10\% & 20\%  \\
    \hline
    Part-A2\citep{shi2021partA2} & 0.04 & 62.3 &  67.1 & 70.9  \\
    PVRCNN++\citep{shi2023pv} & 40.0 & 66.0 &  70.1 & 70.9  \\
    Train from scratch & 41.7 & 67.2&  69.6&  70.8  \\
    \textbf{\systemname(Ours)} & \textbf{50.4}\scriptsize{(+8.7)} & \textbf{70.0}\scriptsize{(+3.0)} & \textbf{70.8}\scriptsize{(+1.3)} & \textbf{71.9}\scriptsize{(+1.1)}\\
    % \bottomrule
    \hline
  \end{tabular}
  
  \label{tab:detection}
\end{table*}

\subsection{Transferring for Semantic Segmentation}
\label{segmentation}

In this section, we conduct experiments to transfer the pre-trained backbone for point cloud semantic segmentation. We compare our method with the state-of-the-art approaches on two evaluation protocols, i.e., linear probing and fine-tuning on point clouds semantic segmentation.

\noindent\textbf{Fine-tuning on Semantic Segmentation:}
We transfer the pre-trained backbone on semantic segmentation by adding a classification head and training them together. We use a linear combination of the cross-entropy and the Lov\'asz Softmax loss~\citep{berman2018lovasz} as our resultant loss. For all the experiments on semantic segmentation, we fine-tune for 100 epochs with a batch size of 16 and the learning rate is 0.02 in nuScenes. We evaluate performance of semantic segmentation on nuScenes-lidarseg\citep{caesar2020nuscenes} and SemanticKITTI\citep{behley2019semantickitti}.
 
We compare Our \systemname with SLidR and a few other methods on the few-shot end-to-end semantic segmentation task and show the results in Tab.~\ref{tab:segmentation}\textcolor{black}{, where 'P-gain' represents the performance improvement of the pretraining method compared to the "Train from Scratch" baseline.} . 
We observe that all representation pre-training methods are better than random initialization. In particular, our method performs the best. The $mIoU$ is  $14.4\%$ and $18.3\%$ higher than random initialization  on nuScenes-lidarseg and SemanticKITTI, and  $4.4\%$  and $5.1\%$ higher than SLidR on the two datasets.

This proves the advantage of exploiting the hybrid-view features in distilling knowledge from pre-trained image networks. We further evaluate the performance of our method in fine-tuning with different percentages of annotated training data: 1\%, 5\%, 10\%, 25\%, and 100\%. The results are shown in Tab.~\ref{tab:percentagesegmentation}. Again, our method performs the best. 
Compared to random initialization, our $mIoU$ improvement is $14.4\%$ for $1\%$ annotation, $10.4\%$ for $5\%$ annotation, $6.3\%$ for $10\%$ annotation, $5.3\%$ for $25\%$ annotation, and $2.4\%$ for $100\%$ annotation. These results verify that our approach does provide a better initialization of point cloud networks.

%-------------------------------------------------------------------------
\begin{table}[t]
  \centering
  \caption{Performance comparison of different pre-training methods for semantic segmentation by linear probing on nuScenes.}
  \addtolength\tabcolsep{3.5pt}
  \begin{tabular}{@{}c|c@{}}

    \hline
    Initialization & Linear Prob(mIoU) \\
    \hline
    Random & 6.8 \\
    
    PointContrast*\citep{xie2020pointcontrast} & 21.9 \\
    DepthContrast*\citep{zhang2021self} & 22.1 \\
    PPKT*\citep{liu2021learning} & 36.4 \\
    SLidR & 38.8 \\
    \textbf{\systemname(Ours)} & \textbf{39.5}\\
    % \bottomrule
    \hline
  \end{tabular}
  \label{tab:linear-prob}
\end{table}

\begin{table}[t]
    \caption{Ablative experiments of different views to perform contrastive distillation when pre-training. We present the $mIoU (\%)$ of semantic segmentation by fine-tuning pre-trained networks using 1\% of the annotated training data in nuScenes. ``Random'' means this model is randomly initialized and trained from scratch. }%The best results are marked with bold font.}
    \addtolength\tabcolsep{6.5pt}
    
  \centering
  \begin{tabular}{cccc}
    % \toprule
    \hline
    Method &  image-plane & BEV & mIoU \\
    % \midrule
    % \hline
    \hline
    % Random &\XSolidBrush &\XSolidBrush & 28.8  \\
    Random &$\times$ &$\times$ & 28.3  \\
    w/o IPV  &$\times$ & \checkmark& 37.8  \\
    w/o BEV  &\checkmark &$\times$ & 41.5  \\
    
    \textbf{\systemname} &\checkmark &\checkmark & \textbf{42.7}  \\ %\textbf{42.2}
    % \bottomrule
    \hline
  \end{tabular}
  
  \label{tab:ablation_1}
\end{table}

\noindent\textbf{Linear Probing:} Next, we evaluate the transferring methods on linear probing. 
 
Here, we add a linear classification head based on the pre-trained 3D backbone. During linear probing, we freeze the backbone and only train the classification head. As in fine-tuning, a linear combination of the cross-entropy and the Lov\'asz-Softmax loss\citep{berman2018lovasz} are utilized as our loss. We train on nuScenes for 50 epochs with a batch size of 16 on 4 GPUs with 100\% of the available annotations. The initial learning rate is 0.05.

The comparison results are summarized in Tab.~\ref{tab:linear-prob}. Among all the methods, ours is again the best performing, with mIoU of $39.5\%$.  This again shows that our hybrid-view distillation can learn more discriminative point cloud representations than single-view distillation such as SLidR\citep{sautier2022image} and PPKT\citep{liu2021learning}.

\subsection{Transferring for Few-Shot Object Detection}
The object detection task aims to recognize foreground objects from raw data, such as point clouds, which is a fundamental problem in computer vision. Generally, it first uses a backbone to learn representations from raw data and then translates the necessary attributes of the object bounding boxes from the learned representations. To verify the effectiveness of our pre-train method, we fine-tune a detection network, PointRCNN~\citep{shi2019pointrcnn}, by loading the parameters of the backbone pre-trained with our approach. 

We conduct our few-shot object detection based on OpenPCDet\citep{openpcdet2020}. In the toolbox, the backbone of PointRCNN is replaced by the pre-trained network SR U-net, and then trained end-to-end for object detection. %We compare the result with random initialization and SLidR to verify the generalization of our method. 
We fine-tune the restructured detection network for 80 epochs on 4 GPUs with a batch size of 16. We use the default settings in OpenPCDet except that the learning rate is changed to 0.01.

\textcolor{black}{
    On the validation set of KITTI dataset, we performed an extensive comparison between our pretrained model and the train-from-scratch baseline, utilizing varying percentages of annotated training data. The results of this comparison are delineated in Tab.~\ref{tab:detection}. Once again, our method stands out as the top performer in this evaluation. It is noteworthy that our achieved Mean Average Precision ($mAP$) surpasses the train-from-scratch with 1\% annotated data by a significant margin of 8.7\%. In addition, we compared our pretrained model to state-of-the-art (SOTA) 3D detection methods. Our pretrained model consistently demonstrated its superiority over the cutting-edge Part-A2 and PVRCNN++ methods.}

\subsection{Ablative Experiments}

\begin{table*}[t]
\caption{Performance comparison of methods for fine-tuning on semantic segmentation with 1\% percent of annotated training data on SemanticKITTI. The results are the $mIoU (\%)$ on the validation set of SemanticKITTI.} %The best results are marked with bold font.}
\centering
\newcommand*\rotext{\multicolumn{1}{R{60}{1em}}}
\setlength{\tabcolsep}{2pt}
% \resizebox{0.8\textwidth}{!}{
\begin{tabular}{c| c c c c c c c c c c c c c c c c c c c |c}
% \toprule
\hline
    Method
    & \rotext{car}
    & \rotext{bicycle}
    & \rotext{motorcycle}
    & \rotext{truck}
    & \rotext{\small{other-vehicle}}
    & \rotext{person}
    & \rotext{bicyclist}
    & \rotext{\small{motorcyclist}}
    & \rotext{road}
    & \rotext{parking}
    & \rotext{sidewalk}
    & \rotext{\small{other-ground}}
    & \rotext{building}
    & \rotext{fence}
    & \rotext{vegetation}
    & \rotext{trunk}
    & \rotext{terrain}
    & \rotext{pole}
    & \rotext{traffic-sign}
    &  \textbf{mIoU}
    % & \rotext{\bf fwIoU}
    \\
% \midrule
\hline
% \hline
    Random   
        & 83.3
        & 0.0
        & 0.0
        & 0.0
        & 0.4
        & 0.0
        & 0.0
        & 0.0
        & 80.0
        & 0.0
        & 58.8
        & 0.0
        & 76.0
        & 27.4
        & 81.3
        & 33.5
        & 65.0
        & 38.8
        & 21.9
        & 29.8
    \\
    % \hline
    \textcolor{black}{w/o IPV  }
        % & \textbf{94.4}
        & \textcolor{black}{91.3}
        & \textcolor{black}{2.8}
        & \textcolor{black}{17.4}
        & \textcolor{black}{20.0}
        & \textcolor{black}{11.9}
        & \textcolor{black}{29.5}
        & \textcolor{black}{28.3}
        & \textcolor{black}{0.0}
        & \textcolor{black}{88.6}
        & \textcolor{black}{25.4}
        & \textcolor{black}{71.4}
        & \textcolor{black}{0.0}
        & \textcolor{black}{86.3}
        & \textcolor{black}{35.5}
        & \textcolor{black}{82.7}
        & \textcolor{black}{57.4}
        & \textcolor{black}{67.5}
        & \textcolor{black}{60.9}
        & \textcolor{black}{43.2}
        & \textcolor{black}{43.2}
    \\
    % \hline
    w/o BEV  
        % & \textbf{94.4}
        & 94.4
        & \textbf{11.7}
        & 24.2
        & 30.8
        & 8.4
        & \textbf{47.6}
        & 45.6
        & 0.0
        & 89.1
        & 24.1
        & 72.8
        & 0.0
        & 85.8
        & 37.9
        & 84.6
        & 59.4
        & \textbf{71.2}
        & \textbf{62.5}
        & \textbf{47.8}
        & 47.3
    \\  
    % \hline
    \textbf{\systemname}  
        & \textbf{94.5}
        % & 93.6
        % & \textbf{12.9}
        & 6.4
        & \textbf{36.8}
        & \textbf{49.3}
        & \textbf{16.2}
        & \textbf{47.4}
        & \textbf{47.5}
        & 0.0
        & \textbf{90.9}
        & \textbf{28.9}
        & \textbf{74.8}
        & 0.1
        & \textbf{88.6}
        & \textbf{46.3}
        & \textbf{85.5}
        & \textbf{62.1}
        & \textbf{71.0}
        & 60.7
        & 45.5
        & \textbf{50.1}\\
% \bottomrule
\hline
\end{tabular}
% }
\label{tab:semanticfinetune1}
\end{table*}

We further conduct ablative experiments to evaluate the design choices in our proposed approach. 

\vspace{4pt}\noindent \textbf{Effect of Different Views: }
% In this experiment, 
We study the effect of different views in our contrastive distillation framework.
We compare three networks: 
(1) train from scratch, (2) \systemname without BEV distillation, \textcolor{black}{(3) \systemname without IPV distillation,} and \textcolor{black}{(4)} \systemname. After pre-training, we transfer the \textcolor{black}{four} backbones on  semantic segmentation with 1\% annotated training data and test them as in Sec.~\ref{segmentation}.

\begin{table}[t]
  \centering
  \caption{
  Ablative experiments of different design choices to build BEV features. We present the mIoU of semantic segmentation by fine-tuning pre-trained networks using 1\% of the annotated training data in nuScenes.}
  \addtolength\tabcolsep{14pt}
  \begin{tabular}{c c}
    \hline
    Method &   mIoU \\
    \hline
    w/o point    & 39.6  \\
    sparse point depth  & 41.2  \\
    Point guidance depth prediction & 42.2 \\
    \textbf{\systemname}  & \textbf{42.7}  \\
    \hline
  \end{tabular}
  \label{tab:ablation_2}
  \vspace{-8pt}
\end{table}

The results are presented in Tab.~\ref{tab:ablation_1}. The IPV-only (w/o BEV in this table) models achieve $41.5\%$ mIoU, improving the randomly initialized one with 13.2\%. \textcolor{black}{The BEV-only (w/o IPV in this table) models achieve $37.8\%$ mIoU, improving the randomly initialized one with 9.5\%.} Our \systemname combines both of BEV and IPV, and further boost the performance to $42.7\%$ mIoU, demonstrating the effectiveness of our hybrid-view distillation framework.

\textcolor{black}{ A detail comparison in different classes for semantic segmentation in SemanticKITTI is presented in Tab.~\ref{tab:semanticfinetune1}. When comparing our \systemname to the w/o BEV model, our approach achieves similar or superior performance in almost all classes, except for bicycle, pole, and traffic-sign. These three object classes are thin and extremely challenging to recognize from the bird's-eye view due to their limited spatial presence and intricate details. The BEV representation may struggle to capture these nuanced characteristics accurately, resulting in decreased recognition and segmentation performance for these classes. However, it's worth noting that our method using BEV-only features outperforms the w/o BEV model in categories such as other-vehicle, building, and parking. This suggests the effectiveness of BEV contrast distillation, particularly for larger objects, where the BEV view excels. The combination of BEV-based and IPV-based contrast distillation contributes to significant overall improvements, demonstrating that BEV and IPV complement each other effectively. This combined approach offers a promising strategy for 2D-to-3D knowledge transfer in our framework.
}

\def\svgwidth{\linewidth}
    \begin{figure*}[th]
    % \vspace{-2pt}
      \centering
      \includegraphics[width=0.9\linewidth]{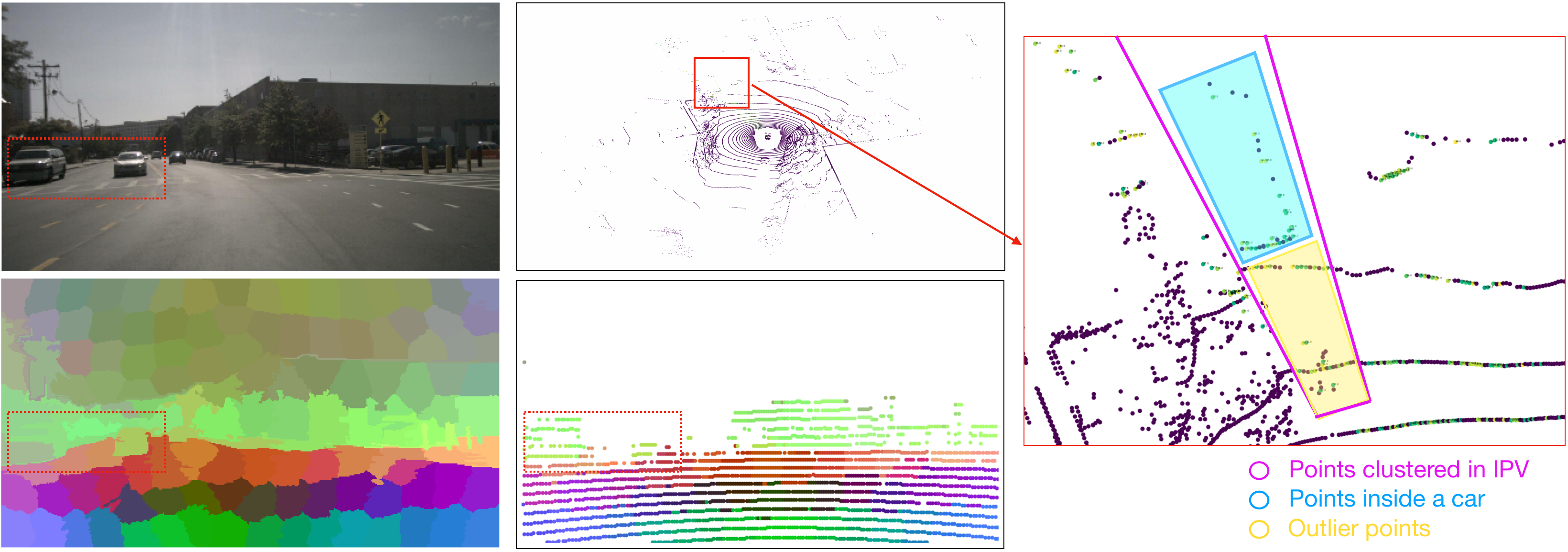}
    \caption{\textcolor{black}{The figure shows the raw image and superpixels on the left, point clouds in BEV view and superpoints on the middle, and The zoomed-out image focuses on the superpoints surrounding the selected car on the right. Points in the red boxes represent the same area. While points in purple are clustered as superpoints to represent the left car in the image, it contains not only the car (points in black area) but also part of ground (points in green area), introducing ambiguity for points.} }
    	\label{fig:bev_better}
    \end{figure*}
    \def\svgwidth{\linewidth}
    
\textcolor{black}{Additionally, we did qualitative analysis in Figure~\ref{fig:bev_better}. The bird-eye view effectively preserves the 3D layout of the scene. In image-plane view distillation, the correspondence between superpoints and superpixels hinges on a projection-based approach. However, this method can introduce a challenge wherein points from varying depths may project to closely situated locations on the image-plane view, potentially leading to the misgrouping of these points into the same cluster, thereby creating ambiguity. As in Figure~\ref{fig:bev_better}, while points in purple area are clustered as superpoints to represent the left car in the image, it contains both the car (points in black area) and part of ground (points in green area). In contrast, the integration of depth information and the re-projection of pixels into 3D space within the bird-eye view representation alleviates this issue, providing a more accurate representation of the scene.}

\textcolor{black}{In addition, the bird-eye view representation offers a mitigation strategy for occlusion and scaling issues, which are notably more complex to resolve in the image-plane view. These inherent advantages are particularly advantageous for the recognition of large objects, as exemplified in Table~\ref{tab:semanticfinetune1}.
     }

\vspace{4pt}\noindent \textbf{Effect of Point Supervision: }

We study the effect of point supervision on our HVDistill and present the results in Table~\ref{tab:ablation_2}.

In this table, ``w/o point'' means predicting the depth distribution for each point without the guidance or the supervision of projected points, ``sparse point depth'' indicates only the pixels with projected points are transformed to the BEV representation with the depth provided by the corresponding point, ``point guided depth distribution'' concatenated the sparse depth map provided by projected points and the feature map to predict the depth distribution without extra supervision, and our method devises the depth module with the supervision of projected points to make depth estimation. Among the competitors, our method with projected point supervision achieves the best performance, highlighting the importance of integrating point-supervised depth prediction in image-to-BEV transformation in our HVDistill.

\begin{figure}[t]
    \centering
    \includegraphics[width=0.5\textwidth]{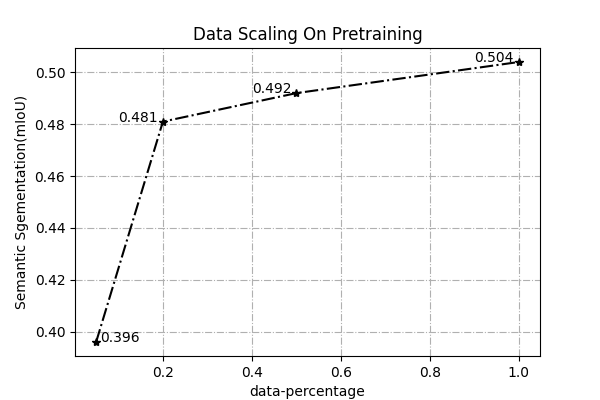}
    \caption{Performance of different training data for semantic segmentation by fine-tuning on SemanticKITTI. We use 1\% training data for fine-tuning.}
    \label{fig:data-scaling}
\end{figure}

\vspace{4pt}\noindent \textbf{Effect of Scaling Up Pretraining Data:}
{In addition to evaluating the performance of \systemname, we conduct an experiment to assess the impact of using different portions of nuScenes. We present the evaluated results of semantic segmentation on nuScenes in Table~\ref{tab:percent-pretrain} and on SemanticKITTI in Figure~\ref{fig:data-scaling}. The nuScenes training data are divided into subsets representing varying sizes. We observe notable variations in performance when utilizing different portions of the pretraining data. Specifically, we find that larger portions of pretraining data consistently yield improved results compared to smaller or limited subsets of these two datasets. These findings suggest that the inclusion of additional pretraining data has the potential to further enhance the performance of our proposed method. With the availability of a larger dataset, we anticipate even more substantial gains in feature extraction.}

\begin{table}[t]
  \centering
  \caption{Performance of different training data for semantic segmentation by fine-tuning on nuScenes. We use 1\% training data for fine-tuning. }
  \addtolength\tabcolsep{12.5pt}
  \begin{tabular}{ccc}
  
    % \toprule
    \hline
    Data for pretraining & mIoU & fwIoU \\
    \hline
    5\% &  35.7 & 72.3 \\
    20\% &  38.9 & 75.0\\
    50\% & 41.2 & 76.7 \\
    100\% & 42.7 &  77.1\\
    % \bottomrule
    \hline
  \end{tabular}
  \label{tab:percent-pretrain}
\end{table}

\begin{table}[t]
  \centering
  \caption{Performance comparison of our method with different image teachers for semantic segmentation by fine-tuning on nuScenes. We use 1\% training data for fine-tuning. }
  \addtolength\tabcolsep{1.5pt}
  \begin{tabular}{@{}c c c c @{}}
  
    % \toprule
    \hline
    Teacher method & top-1 acc(ImageNet) & mIoU & fwIoU \\
    % \midrule
    % \hline
    \hline
    % MoCo & - & & \\
    Dino & 75.3 & 41.4 & 77.0\\
    Supervised & 79.3 &42.9 & 77.2 \\
    MoCoV2 & 71.1 & 42.7 &  77.1\\
    % \bottomrule
    \hline
  \end{tabular}
  \label{tab:teacher-choice}
\end{table}

 \def\svgwidth{\linewidth}

\vspace{4pt}\noindent \textbf{The Effect of Pretrained Image Teacher: }
{To explore the influence of different image teachers on our proposed method, we conduct experiments utilizing multiple pre-trained image teachers in addition to the MoCoV2. The results are presented in Table~\ref{tab:teacher-choice}. Surprisingly, Despite DINO's~\citep{caron2021emerging} superior performance in image classification, this image teacher network is not optimally suited for our \systemname. These results suggest that the choice of pre-trained image teacher should be carefully considered, taking into account its compatibility with the target task. While some image teachers may excel in certain domains, their performance may not necessarily translate directly to improve results in other tasks.}

\vspace{4pt}\noindent \textbf{Comparison of image-plane-view Method: }
{We explore pixel-to-point and superpixel-to-superpoint contrast learning methods on IPV branch in Table~\ref{tab:IPV-choice}. The pixel-to-point contrast learning involves computing at individual pixel and corresponding point, while superpixel-to-superpoint contrast learning aggregates visually similar pixels and translates knowledge in superpixel level. Our results show that the pixel-to-point does yield a lower performance, as the noise of calibration of LiDAR and camera led to suboptimal results. }

\begin{table}[t]
  \centering
  \caption{Performance comparison of different contrast learning methods on IPV branch. We use 1\% training data of nuScenes for fine-tuning on semantic segmentation. }
  \addtolength\tabcolsep{4.5pt}
  \begin{tabular}{c c c}
  
    % \toprule
    \hline
    contrast method on IPV branch & mIoU & fwIoU \\
    % \midrule
    \hline
    % \hline
    None  & 33.4 & 71.8\\
    pixel-to-point  &41.1 & 76.8 \\
    Superpixel-to-superpoint  & 42.7 &  77.1\\
    % \bottomrule
    \hline
  \end{tabular}
  \label{tab:IPV-choice}
\end{table}

\textcolor{black}{\vspace{4pt}\noindent \textbf{Ablation study of $\alpha$ and $\beta$: }{In Tab.4, we present the mIoU scores achieved with different weight ratios, specifically $\alpha:\beta$. We varied these ratios while keeping $\beta$ constant and observed the corresponding performance changes. The table highlights the following findings:
        When $\alpha:\beta$ is set to 1:1, the performance reaches its lowest point, indicating that an equal contribution from both image-plane view and bird's-eye view distillation has a minimal impact on performance.       
        The best performance is achieved when $\alpha:\beta$ is set to 4:1, underscoring the significance of the bird's-eye view distillation in enhancing semantic segmentation.
        We note that while there is variation in performance across different weight settings, the impact on overall performance is relatively minimal.
        }}
    \begin{table}[t]
        \centering
        \caption{Ablation of $\alpha$ and $\beta$. $\alpha:\beta$ = 4:1 is the best choice. }
        \addtolength\tabcolsep{10.5pt}
        \begin{tabular}{c |c}
        \hline
        $\alpha$ : $ \beta$ & mIoU \\
        \hline
        4 : 1 & 50.1 \\
        2 : 1 & 49.8 \\
        1 : 1 & 49.4 \\
        1 : 2 & 49.5 \\
        \hline
    \end{tabular}
    \label{tab:ratio}
\end{table}
        
\subsection{\textcolor{black}{Qualitative Results}}
\begin{figure}[t]
% \vspace{-2pt}
\centering

\begin{minipage}{0.3\linewidth}
\centerline{\small{(a) Sparse point}}
        \includegraphics[width=\textwidth]{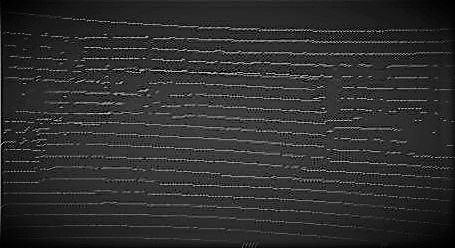}
    \end{minipage}%
    % \quad 
    \hfill
\begin{minipage}{0.3\linewidth}
\centerline{\small{(b) Predicted Depth}}
    % \centering
    \includegraphics[width=\textwidth]{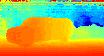}
\end{minipage}
\quad
% \hfill
\begin{minipage}{0.3\linewidth}
\centerline{\small{(c) Image}}
    % \centering
    \includegraphics[width=\textwidth]{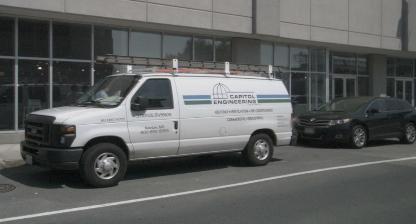}
    % \centerline{ }
\end{minipage}
\\[\smallskipamount]
 
\begin{minipage}{0.3\linewidth}
        \includegraphics[width=\textwidth]{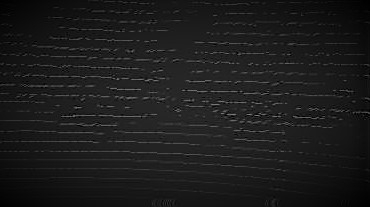}
    \end{minipage}%
    % \quad 
    \hfill
\begin{minipage}{0.3\linewidth}
    \includegraphics[width=\textwidth]{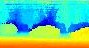}
\end{minipage}
\quad
\begin{minipage}{0.3\linewidth}
    % \centering
    \includegraphics[width=\textwidth]{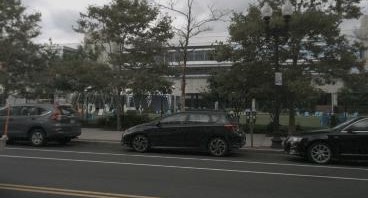}
    % \centerline{ }
\end{minipage}
\\[\smallskipamount]

\begin{minipage}{0.3\linewidth}
        \includegraphics[width=\textwidth]{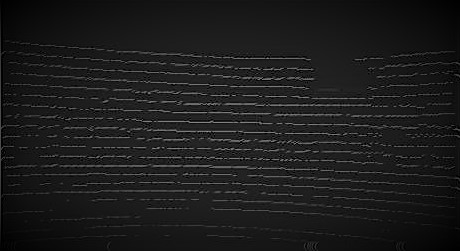}

    \end{minipage}%
    % \quad 
    \hfill
\begin{minipage}{0.3\linewidth}
    % \centering
    \includegraphics[width=\textwidth]{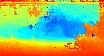}
\end{minipage}
\quad
\begin{minipage}{0.3\linewidth}
    % \centering
    \includegraphics[width=\textwidth]{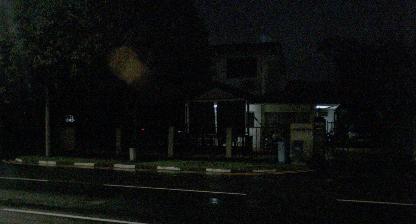}
    % \centerline{ }
\end{minipage}

\caption{Visualization of the predicted depth of different objects. With sparse point cloud supervision, image features can predict the dense depth and preserve geometric information.  Each row represents a specific scene.}
\label{fig:visualization_depth}
\end{figure}

\vspace{4pt}\noindent \textbf{\textcolor{black}{Visualization of predicted depth:} }{We visualize the discrete depth predictions obtained from our model in Figure~\ref{fig:visualization_depth}, which represent the depth value with the maximum probability for each pixel. We have also included a comparison with sparse depth information derived from point clouds. The results demonstrate that our model performs well in diverse scenes, including both daytime and nighttime scenarios, and can accurately predict depth for a wide range of objects, such as cars and trees. Overall, our approach shows promising results for depth estimation in real-world settings.}

\def\svgwidth{\linewidth}
\def\svgwidth{\linewidth}

\begin{figure*}[th]
	\centering
 % \caption{\\3}
	
	\begin{minipage}[t]{0.24\linewidth}
		\centering
		\includegraphics[width=\linewidth]{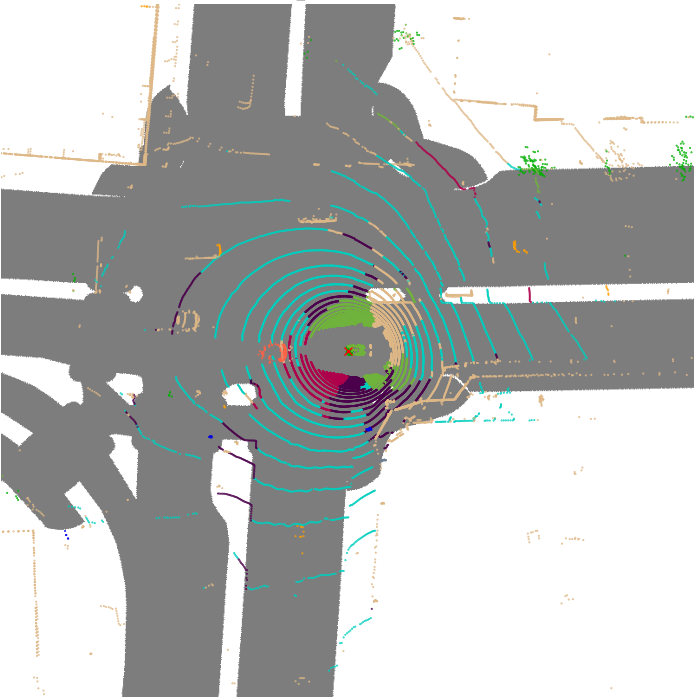} \\
        \includegraphics[width=\linewidth]{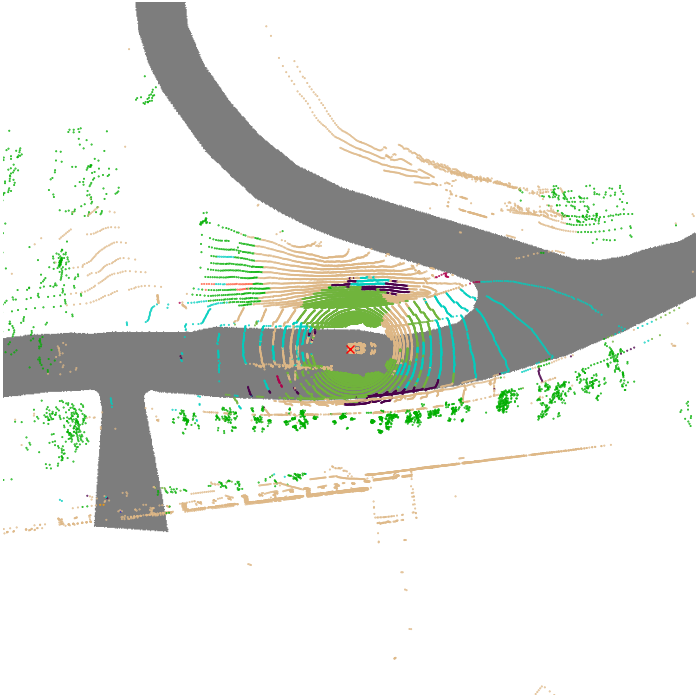}\\
        \includegraphics[width=\linewidth]{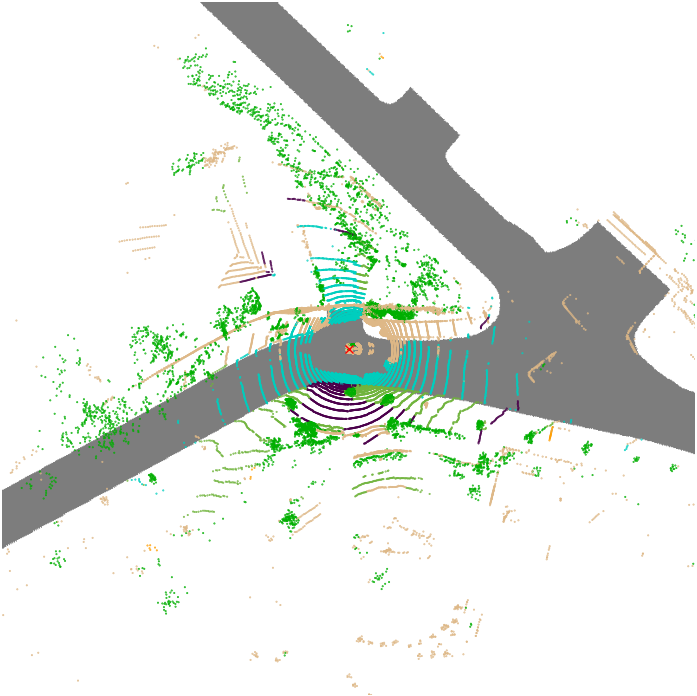} 
        \subcaption{Scratch}
	\end{minipage}
	\begin{minipage}[t]{0.24\linewidth}
		\centering
		\includegraphics[width=\linewidth]{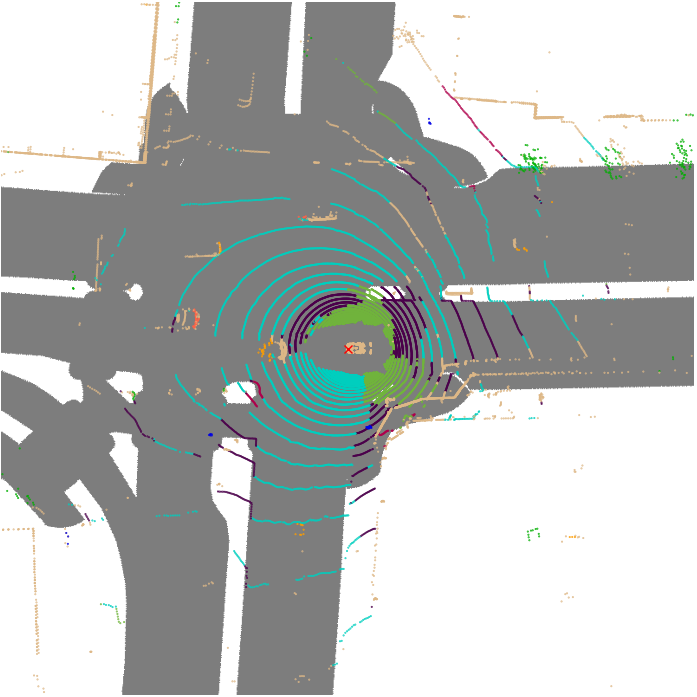} \\
        \includegraphics[width=\linewidth]{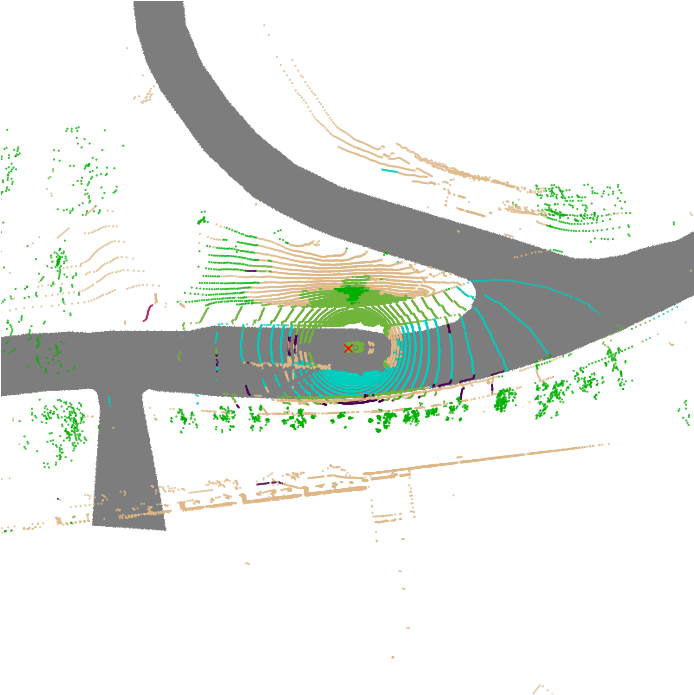}\\
        \includegraphics[width=\linewidth]{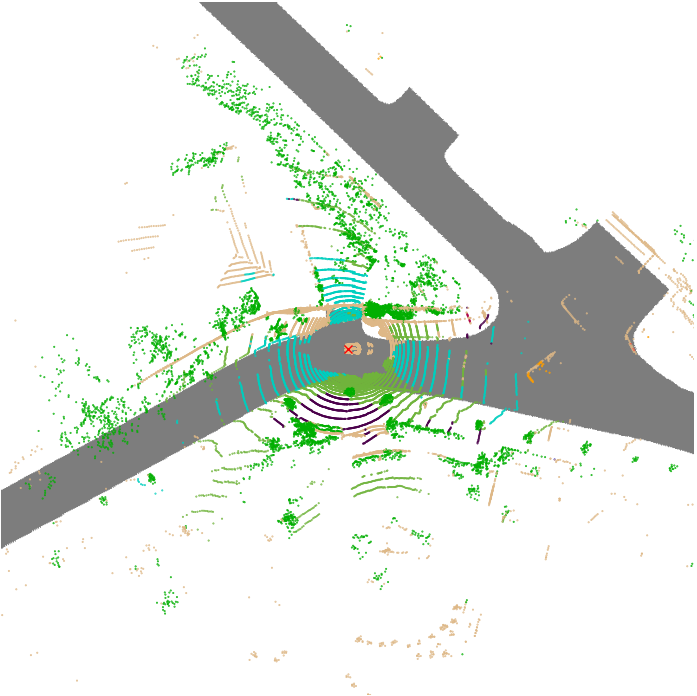} 
        % \caption{\\3}
        \subcaption{SLiDR}
	\end{minipage}
 \begin{minipage}[t]{0.24\linewidth}
		\centering
		\includegraphics[width=\linewidth]{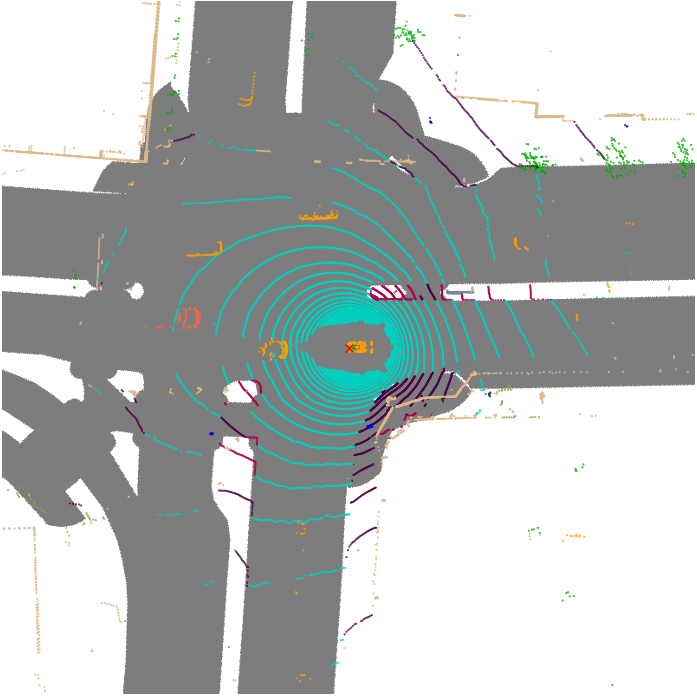} \\
        \includegraphics[width=\linewidth]{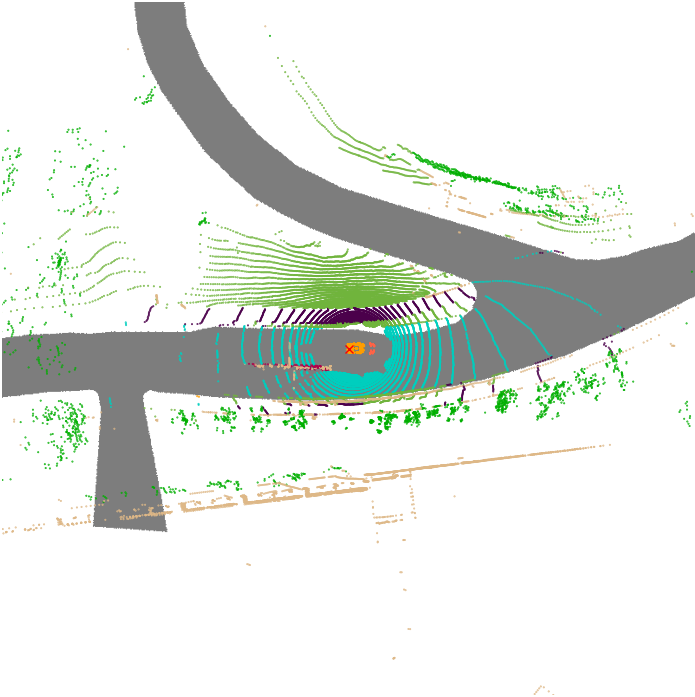}\\
        \includegraphics[width=\linewidth]{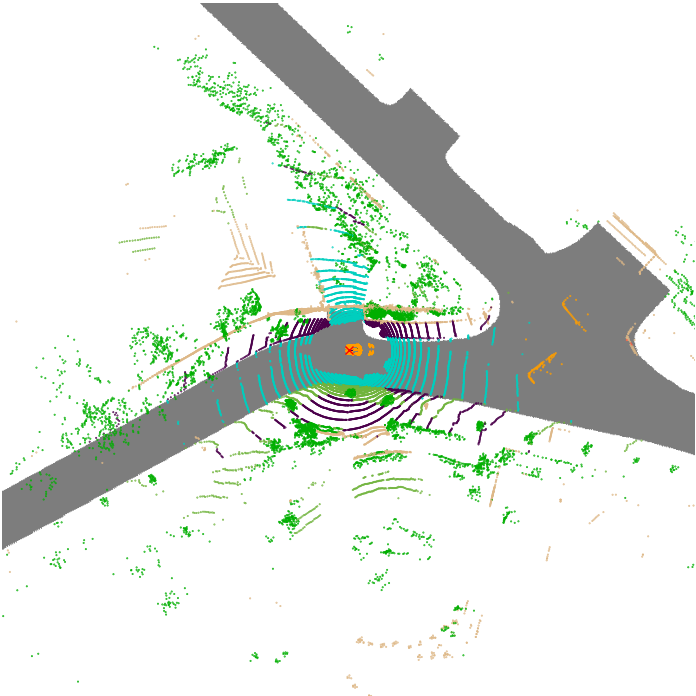} 
        % \caption{\\3}
        \subcaption{Ours}
	\end{minipage}
 \begin{minipage}[t]{0.24\linewidth}
		\centering
		\includegraphics[width=\linewidth]{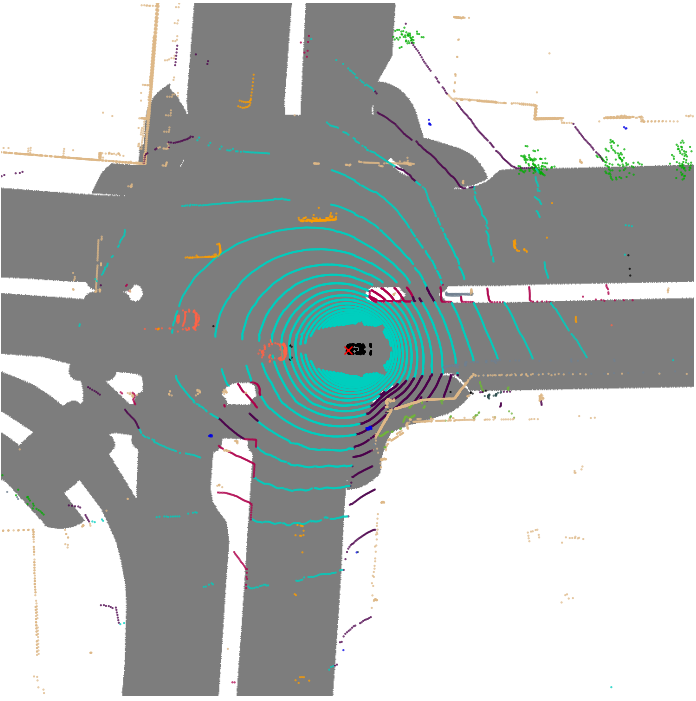} \\
        \includegraphics[width=\linewidth]{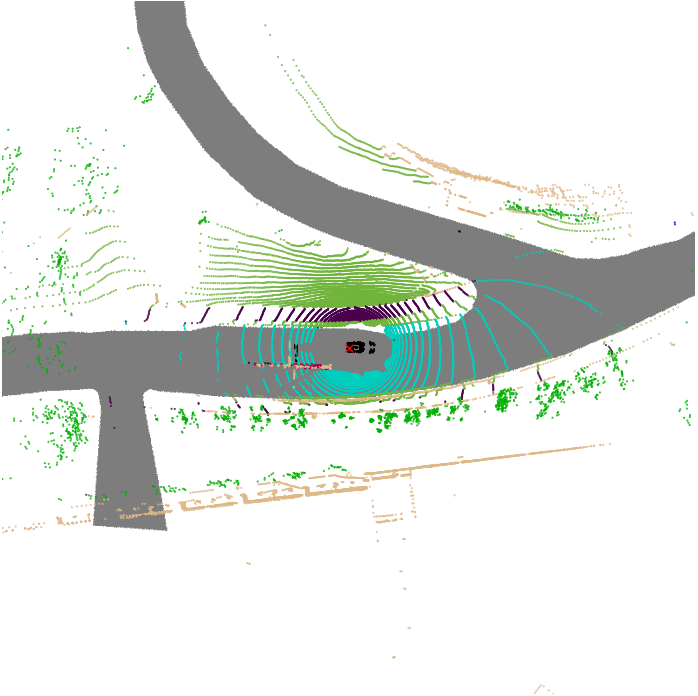} \\
        \includegraphics[width=\linewidth]{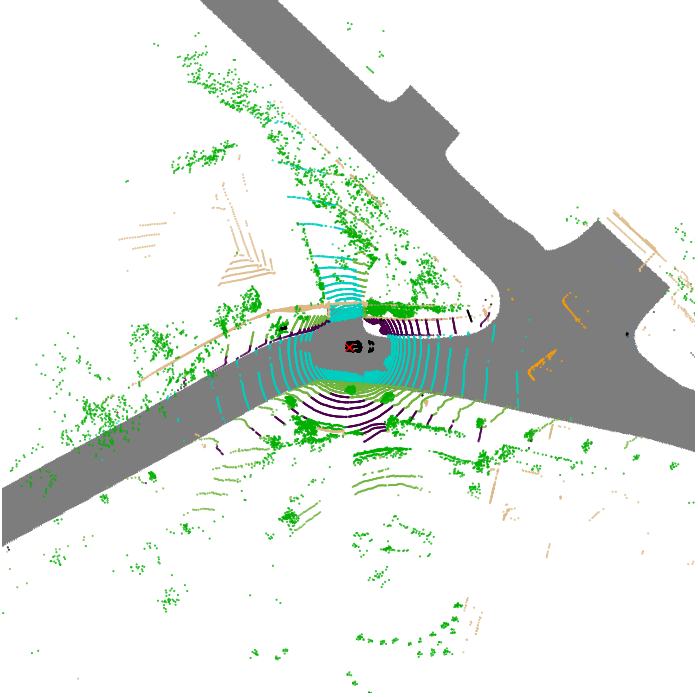} 
        \subcaption{Ground Truth}
        
	\end{minipage}
 \caption{Qualitative results of 3D semantic segmentation on the nuScenes validation set. }

\label{fig:qualitative} 
\end{figure*}

\vspace{4pt}\noindent \textcolor{black}{\textbf{Visualization of Semantic segmentation results:} Figure~\ref{fig:qualitative} illustrates the semantic segmentation results on the nuScenes validation set, including (a) Ground Truth, (b) Train from scratch, (c) SLiDR, and (d) Our method. Our model can predict accurate class for distant and highly occluded objects, demonstrating the high-quality predictions of our model.}

\section{Conclusion}
\label{conclusion}
In this paper, we propose \systemname, a self-supervised 2D-to-3D distillation method  to transfer knowledge from a pre-trained image model to a point cloud  neural network via two cross-modality contrastive losses, one from the image plane view (IPV) and the other from BEV. The BEV contrastive distillation utilizes the depth information in the point clouds  to preserve the geometric information, providing a complement for the mainstream IPV contrastive distillation.  The experiments demonstrate that our pre-train method brings significant improvement in downstream tasks compared to the state-of-the-art. We believe that our method takes a step towards more effective knowledge transfer between modalities.  %We hope our approach can inspire future research on knowledge transferring between a variety of sensors. 

\parsection{Data availability} 
The nuScenes~\citep{caesar2020nuscenes} dataset and nuScenes-lidarseg~\citep{caesar2020nuscenes} dataset can be obtained from \url{https://www.nuscenes.org/}. The KITTI~\citep{geiger2012we} dataset and SemanticKITTI~\citep{behley2019semantickitti} dataset can be obtained from \url{https://www.cvlibs.net/datasets/kitti/}. The code that supports the findings of this study are available from the corresponding author, Yanyong Zhang, upon reasonable request.

\section*{Declarations}

\textbf{Conflict of interest} There are no conflicts to declare.

\bibliographystyle{spbasic}

\bibliography{main}

\end{document}